\documentclass[conference]{IEEEtran}
\IEEEoverridecommandlockouts
\pdfoutput=1
\usepackage[T1]{fontenc}
\usepackage{mathptmx}

\usepackage{graphicx}
\newcommand{\dv}[1]{\raisebox{-4.5pt}[7.5pt][2.5pt]{\includegraphics[page=#1]{dev-snippets.pdf}}}
\usepackage{booktabs}
\usepackage{multirow}
\usepackage{xcolor}
\PassOptionsToPackage{hyphens}{url}
\usepackage[hidelinks]{hyperref}
\usepackage{url}

\newcommand{\pmsd}[1]{{\scriptsize$\pm$#1}}
\newcommand{\hfid}[1]{\texttt{\small #1}}

\begin{document}

\title{NEPSEE: A Double-Annotated Benchmark for Nepali Sentiment Analysis}

\author{\IEEEauthorblockN{Gaurab Baral}
\IEEEauthorblockA{\textit{University of Cincinnati}\\
Cincinnati, Ohio, USA\\
\textit{baralgb@uc.edu}}
\and
\IEEEauthorblockN{Mimansak Nepal}
\IEEEauthorblockA{\textit{Tribhuvan University}\\
Kirtipur, Kathmandu, Nepal\\
\textit{mimansaknepal@gmail.com}}
\and
\IEEEauthorblockN{Grishma Basnet}
\IEEEauthorblockA{\textit{Subhakamana English School}\\
Urlabari, Morang, Nepal\\
\textit{grishmabsnt@proton.me}}}

\maketitle

\begin{abstract}
We present NEPSEE, a benchmark for Nepali sentiment analysis with 3,189 Devanagari sentences from tweets of Nepali public figures. Two annotators labelled each sentence blind on a five-point scale, and we keep the sentences where they agree on polarity. We evaluate 12 fine-tuned encoder configurations, TF-IDF baselines and five zero-shot large language models (LLMs), three commercial and two with open weights. On three classes, NepBERTa with a fixed tokenizer gets a macro-F1 of 0.877, IndicBERTv2 0.867 and the best TF-IDF model 0.766. Three commercial LLMs get 0.882 to 0.896 with no training data, but they are not significantly better than the best encoders. On five points, Claude Opus 5.5 gets 0.838 and beats every encoder by at least 0.10, because the encoders see few extreme examples. A tokenizer audit adds a second result. The default tokenizers of three Nepali BERT models delete the non-spacing Devanagari vowel signs and the virama. When we restore the marks, macro-F1 increases by 0.084 for NepBERTa but decreases by 0.095 for a model pretrained on stripped text. A masked-language-model probe that needs no labels predicts the direction of each effect.
\end{abstract}


\section{Introduction}

Nepali sentiment classification has no shared, fine-grained benchmark. Published Nepali sentiment datasets label whole tweets, comments or posts with two or three classes~\cite{sitaula2021deep,singh2020nepsa,rauniyar2023naet,sitoula2025nepemo}. The Nepali language-understanding benchmarks have no sentiment task~\cite{timilsina2022nepberta} or a binary task built partly from translated English data~\cite{nyachhyon2025nlue}. Their label sets, domains and metrics differ, so their scores do not compare.

A benchmark score also depends on who made the labels. Models learn the conventions of the annotators of their training set~\cite{geva2019modeling}. Thus a high score can show agreement with one person and not with other readers~\cite{tedeschi2023superhuman}. To prevent this problem, two annotators labelled our sentences blind, and we keep only the labels that they agree on. Their agreement is low on the five-point scale (unweighted $\kappa=0.289$), but the two persons give opposite polarity on only 2.4\% of sentences.

A second problem is in the tokenizers. Devanagari writes most vowels as combining marks attached to consonants, and it joins consonants with a virama. The three Nepali BERT checkpoints on the Hugging Face Hub have no tokenizer configuration file. Thus the library uses default BERT settings, which delete every Unicode non-spacing mark. These marks include the vowel signs \dv{1}, \dv{2}, \dv{3}, \dv{4} and the virama \dv{5}.

We use a new dataset\footnote{\url{https://github.com/Nyvora-Vision-Labs/NEPSEE-A-Nepali-Sentiment-Emotion-and-Explanation-Dataset}} and make four contributions:
\begin{itemize}
\item a benchmark of 3,189 sentences with the agreed polarity of two blind annotators;
\item a five-point task with two acceptable labels for each sentence where the annotators differ only in intensity;
\item results for 12 encoder configurations, TF-IDF baselines, and three commercial and two open-weight LLMs on the same test sentences;
\item an audit of the Nepali BERT tokenizers, with a masked-language-model (MLM) probe that predicts the effect of the fix on each checkpoint.
\end{itemize}

\section{Related Work}

\textbf{Nepali sentiment data} are small and coarse. NepCOV19Tweets has about 33,247 COVID-19 tweets in three classes~\cite{sitaula2021deep}, and later work used hybrid features and CNNs on it~\cite{shahi2022hybrid,sitaula2023multichannel}. NepSA labels 3,068 YouTube comments with binary polarity for six aspects~\cite{singh2020nepsa}, and earlier work used classical classifiers and lexicons~\cite{tamrakar2020absa,piryani2020nepali}. NAET labels 4,445 election tweets in three classes, and its best sentiment baseline gets a macro-F1 of 0.467~\cite{rauniyar2023naet}. NepEMO labels 4,462 Reddit posts~\cite{sitoula2025nepemo}, and a 2026 shared task covers Nepali memes~\cite{acharya2026teamherald}. Other resources target hate speech and offensive language~\cite{thapa2023nehate,niraula2021offensive,thapa2025chipsal}, and none uses a five-point scale.

\textbf{Nepali encoders.} NepBERTa is a BERT-base model pretrained on about 0.8 billion words of Nepali news, and its Nep-gLUE benchmark has no sentiment task~\cite{timilsina2022nepberta}. NLUE adds a binary sentiment task~\cite{nyachhyon2025nlue}, and Maskey et al.\ compare several Nepali encoders~\cite{maskey2022nepali}. We found no study of the tokenizer settings of these checkpoints. Fine-grained sentiment has a longer history in English~\cite{socher2013sst,nakov2016semeval,rosenthal2017semeval}.

\textbf{Agreement and label variation.} Krippendorff's $\alpha$ accepts ordinal scales, and $\alpha=0.667$ is the usual threshold for tentative conclusions~\cite{krippendorff2011computing,artstein2008survey}. Human label variation can be signal and not only noise~\cite{plank2022problem}, and soft labels are one way to train with it~\cite{uma2021learning}. Zero-shot LLMs can label text more accurately than crowd workers~\cite{gilardi2023chatgpt}, but they do not beat the best fine-tuned models on most classification tasks~\cite{ziems2024css}.

\section{Data}

\subsection{Corpus}
We collected tweets from public accounts of Nepali politicians and journalists. We kept the tweets that contain Devanagari and removed Latin script, URLs and Latin hashtags. Then we removed tweets with fewer than four Nepali words. We split each tweet at the danda (\dv{6}) and at line breaks, and we removed sentences shorter than five words. The dataset has 5,614 sentences from 3,388 tweets of 11 accounts (9 politicians and 2 journalists), posted between December 2017 and August 2026.

\subsection{Blind double annotation}
Two annotators, A and B, labelled all 5,614 sentences. Each sentence gets one of six labels: strongly negative (SNeg), negative (Neg), neutral (Neu), positive (Pos), strongly positive (SPos) or ``don't know''. The brief tells annotators to judge the sentiment that the writer expresses, not the event. Annotators use a mobile web application, which saves each label to a hosted database immediately. Each annotator has a personal login, and the application shows only the labels of that annotator. No annotator saw the labels of another person or of an LLM.

\subsection{Agreement between the annotators}
Table~\ref{tab:agree} gives the agreement on the 4,943 sentences where both annotators give a polar label. They choose the same five-point label on 49.5\% of sentences, and unweighted $\kappa$ is only 0.289. But 95.6\% of their label pairs are at most one step apart, and ordinal $\alpha$ is 0.625. They give opposite polarity on only 2.4\% of sentences.

\begin{table}[t]
\centering
\caption{Agreement between annotators A and B on the 4,943 sentences with two polar labels. Five points: ordinal $\alpha$; three classes: nominal $\alpha$. Brackets: 95\% bootstrap CI (1,000 resamples).}
\label{tab:agree}
\setlength{\tabcolsep}{4pt}
\begin{tabular}{lcccc}
\toprule
Scale & Exact \% & Cohen $\kappa$ & QWK & Krippendorff $\alpha$ \\
\midrule
Five points & 49.5 & 0.289 [.27, .31] & 0.627 & 0.625 [.61, .64] \\
Three classes & 65.5 & 0.465 [.45, .48] & -- & 0.455 [.43, .47] \\
\bottomrule
\end{tabular}
\end{table}

Most disagreements concern the boundary between Neu and a polar label. Annotator B uses Neu for 45.0\% of polar pairs, and annotator A for 29.5\%. Of the 1,934 sentences that annotator A calls Pos, annotator B calls 906 Neu (47\%). Annotator A also uses SPos more often (13.1\% against 3.7\%), and annotator B uses SNeg more often (4.3\% against 0.7\%). Thus the two persons have a different calibration, but they rarely disagree on the direction of the sentiment.

\subsection{The agreed benchmark}
We removed 671 sentences where one annotator chose ``don't know'' and 1,704 sentences with a different polarity. Most of the 1,704 have Neu from one annotator and a polar label from the other. Then we removed 50 exact duplicates and kept the first copy. This leaves 3,189 sentences from 2,381 tweets (Table~\ref{tab:data}).

The three-class label is the agreed polarity. On the five-point scale, the two annotators give the same label on 2,423 sentences. The other 766 sentences have two adjacent labels with the same polarity, Neg and SNeg (139) or Pos and SPos (627). For these sentences, we accept both labels as correct. We split the data by tweet with a stratified group 20-fold split (14, 3 and 3 folds), so no tweet crosses two splits.

\begin{table}[t]
\centering
\caption{Labels in each split. Rows with two labels: the annotators differ only in intensity, and both labels are correct. The splits keep each tweet together (1,668, 357 and 356 tweets).}
\label{tab:data}
\setlength{\tabcolsep}{5pt}
\begin{tabular}{lrrrrr}
\toprule
Five-point label & Train & Dev & Test & Total & 3-class \\
\midrule
SNeg & 14 & 1 & 7 & 22 & \multirow{3}{*}{Neg} \\
SNeg or Neg & 106 & 22 & 11 & 139 & \\
Neg & 323 & 58 & 66 & 447 & \\
\midrule
Neu & 710 & 161 & 167 & 1,038 & Neu \\
\midrule
Pos & 602 & 127 & 124 & 853 & \multirow{3}{*}{Pos} \\
Pos or SPos & 437 & 97 & 93 & 627 & \\
SPos & 43 & 9 & 11 & 63 & \\
\midrule
Total & 2,235 & 475 & 479 & 3,189 & \\
\bottomrule
\end{tabular}
\end{table}

\section{Method}

\subsection{Models and baselines}
We fine-tune nine public checkpoints. The multilingual checkpoints are multilingual BERT (\hfid{bert-base-multilingual-cased})~\cite{devlin2019bert}, XLM-R base~\cite{conneau2020xlmr}, MuRIL base~\cite{khanuja2021muril} and IndicBERTv2 (\hfid{IndicBERTv2-MLM-only})~\cite{doddapaneni2023indicxtreme}. The Nepali checkpoints are NepBERTa~\cite{timilsina2022nepberta}, two community Nepali BERTs (\hfid{Rajan/NepaliBERT} and \hfid{Shushant/nepaliBERT}), a Nepali RoBERTa (\hfid{IRIIS-RESEARCH/RoBERTa\_Nepali\_125M}) and a Nepali DeBERTa (\hfid{Sakonii/deberta-base-nepali}). The original IndicBERT~\cite{kakwani2020indicnlpsuite} is gated, so we use IndicBERTv2. We converted the TensorFlow weights of NepBERTa to PyTorch (202 of 202 tensors). Each model has the same head: the first-token hidden state, dropout 0.1 and a linear layer.

The baselines are logistic regression and a linear SVM on TF-IDF features of word 1--2-grams, character 2--5-grams or both. We tune the regularization and class weights on development macro-F1. A majority-class model gives the floor.

\subsection{Tokenizer audit and MLM probe}
The three Nepali BERTs (NepBERTa, Rajan, Shushant) have no \texttt{tokenizer\_config.json}. Thus the \texttt{transformers} library loads them with \texttt{do\_lower\_case=True}, which also strips accents. It decomposes the text and deletes all Unicode non-spacing marks (category Mn). The spacing vowel signs such as \dv{7} and \dv{8} stay. For the sentence \dv{9} \dv{10} \dv{11} \dv{12} \dv{13} (``Nepal's politics is extremely disappointing.''), the NepBERTa tokenizer gives these subwords:
\begin{center}\small
\begin{tabular}{@{}ll@{}}
default: & \dv{14}\\
fixed: & \dv{15}
\end{tabular}
\end{center}
By default, \dv{9} (``of Nepal'') loses its vowel sign, and \dv{11} (``extremely'') loses two viramas and a vowel sign.

We run each Nepali BERT with the default tokenizer and with a fixed tokenizer (\texttt{do\_lower\_case=False}, \texttt{strip\_accents=False}), with the same seeds. For each tokenizer we measure fertility (subwords per word), the share of combining marks lost, and the share of words kept intact. The MLM probe masks 15\% of the tokens of 600 benchmark sentences. It then measures the top-1 accuracy of the MLM head of the checkpoint under each setting. A checkpoint should predict masked tokens best with its pretraining normalization.

\subsection{Training with two acceptable labels}
We train a separate model for each scale. On three classes, the target is the agreed polarity. On five classes, the target is the vote distribution of the two annotators~\cite{uma2021learning}. Thus a sentence with Pos and SPos has the target 0.5 for each label, and a sentence with one label has the target 1.0. The TF-IDF baselines use the same targets: each sentence with two labels gives two training rows with weight 0.5.

We fine-tune with AdamW (learning rate $2\times10^{-5}$, weight decay 0.01, 10\% linear warm-up), batch size 16, maximum length 128 tokens and gradient clipping at 1.0. Training stops after at most six epochs, with early stopping on development macro-F1 (patience 2). Each configuration runs with seeds 1, 2 and 3 on one NVIDIA A100 GPU.

\subsection{Evaluation}
The main metric is macro-F1. On five classes, the \emph{lenient} score counts a prediction as correct if it matches either label of the sentence. We also give a \emph{strict} score on the 375 test sentences where the two annotators chose the same five-point label. Other metrics are accuracy, quadratic weighted kappa (QWK) and the severe-error rate. A severe error has a predicted polarity opposite to the reference polarity. Paired bootstrap tests resample the test set 2,000 times and use the mean macro-F1 over three seeds.

\subsection{Zero-shot LLMs}
\label{sec:llm}
We test three commercial LLMs and two LLMs with open weights. Four LLMs labelled each sentence through OpenRouter on 1 October 2026: Gemini 3.8 Flash, Gemini 3.1 Pro (preview), and the open-weight Qwen3.5-397B-A17B and Qwen3.5-9B. The prompt repeats the annotator brief and the six labels, and asks for JSON output at temperature 0, one sentence per call. The Gemini models used reasoning effort ``low''. The Qwen models used no reasoning, because it gave thousands of tokens per sentence.

Claude Opus 5.5 (\hfid{claude-opus-5-5}) labelled each sentence on 2 October 2026 through the Claude Code agent. Twelve parallel sub-sessions read the sentences in batches of 40 to 60 and wrote one label for each sentence. Unlike the other LLMs, Claude saw the other sentences of its batch. No LLM saw a human label. We count a ``don't know'' or missing answer as an error.

\section{Results}

\begin{table*}[t]
\centering
\caption{Test results on 479 sentences (encoders: mean {\scriptsize$\pm$} s.d.\ over three seeds, in order of three-class macro-F1). Five-point lenient: a prediction is correct if it matches either annotator. Strict: the 375 test sentences with one five-point label. Severe: predicted polarity opposite to the reference. For LLMs, QWK and Severe use only the sentences with a polar answer. Bold: best in each column.}
\label{tab:main}
\setlength{\tabcolsep}{3.5pt}
\begin{tabular}{llcccccccc}
\toprule
& & \multicolumn{2}{c}{Three classes} & \multicolumn{4}{c}{Five points, lenient} & \multicolumn{2}{c}{Five points, strict} \\
\cmidrule(lr){3-4}\cmidrule(lr){5-8}\cmidrule(lr){9-10}
& Model & Macro-F1 & Accuracy & Macro-F1 & Accuracy & QWK & Severe \% $\downarrow$ & Macro-F1 & Accuracy \\
\midrule
\multirow{2}{*}{Baselines}
& Majority class & 0.215 & 0.476 & 0.125 & 0.453 & -- & -- & -- & -- \\
& TF-IDF (best dev) & 0.766 & 0.793 & 0.633 & 0.729 & 0.714 & 5.2 & -- & -- \\
\midrule
\multirow{12}{*}{\shortstack[l]{Fine-tuned\\encoders}}
& NepBERTa (fixed tok.) & 0.877\pmsd{.018} & \textbf{0.891}\pmsd{.016} & 0.721\pmsd{.043} & \textbf{0.863}\pmsd{.024} & 0.897\pmsd{.026} & 0.7\pmsd{0.5} & 0.638\pmsd{.031} & \textbf{0.830}\pmsd{.031} \\
& IndicBERTv2 & 0.867\pmsd{.003} & 0.881\pmsd{.004} & 0.734\pmsd{.001} & 0.859\pmsd{.001} & 0.883\pmsd{.002} & 1.3\pmsd{0.2} & 0.640\pmsd{.004} & 0.826\pmsd{.002} \\
& DeBERTa-base Nepali & 0.858\pmsd{.016} & 0.875\pmsd{.013} & 0.704\pmsd{.031} & 0.842\pmsd{.013} & 0.867\pmsd{.017} & 2.1\pmsd{0.9} & 0.608\pmsd{.024} & 0.808\pmsd{.014} \\
& RoBERTa-Nepali 125M & 0.847\pmsd{.003} & 0.862\pmsd{.003} & 0.703\pmsd{.027} & 0.825\pmsd{.007} & 0.844\pmsd{.004} & 2.6\pmsd{0.5} & 0.607\pmsd{.017} & 0.781\pmsd{.009} \\
& XLM-R base & 0.840\pmsd{.002} & 0.859\pmsd{.002} & 0.661\pmsd{.016} & 0.823\pmsd{.021} & 0.837\pmsd{.014} & 2.5\pmsd{0.0} & 0.592\pmsd{.025} & 0.785\pmsd{.026} \\
& MuRIL base & 0.831\pmsd{.006} & 0.850\pmsd{.008} & 0.487\pmsd{.003} & 0.821\pmsd{.001} & 0.786\pmsd{.006} & 2.6\pmsd{0.3} & 0.470\pmsd{.003} & 0.781\pmsd{.003} \\
& Shushant nepaliBERT (default tok.) & 0.817\pmsd{.005} & 0.845\pmsd{.004} & 0.670\pmsd{.036} & 0.802\pmsd{.020} & 0.818\pmsd{.022} & 3.0\pmsd{0.4} & 0.559\pmsd{.010} & 0.761\pmsd{.027} \\
& NepBERTa (default tok.) & 0.793\pmsd{.015} & 0.817\pmsd{.010} & 0.671\pmsd{.033} & 0.786\pmsd{.003} & 0.805\pmsd{.012} & 3.8\pmsd{0.2} & 0.575\pmsd{.023} & 0.737\pmsd{.006} \\
& Rajan NepaliBERT (fixed tok.) & 0.788\pmsd{.014} & 0.811\pmsd{.013} & 0.618\pmsd{.012} & 0.784\pmsd{.005} & 0.774\pmsd{.014} & 2.5\pmsd{0.4} & 0.546\pmsd{.016} & 0.744\pmsd{.009} \\
& mBERT (cased) & 0.771\pmsd{.026} & 0.800\pmsd{.026} & 0.630\pmsd{.013} & 0.775\pmsd{.020} & 0.764\pmsd{.017} & 4.7\pmsd{0.5} & 0.555\pmsd{.020} & 0.737\pmsd{.021} \\
& Rajan NepaliBERT (default tok.) & 0.739\pmsd{.011} & 0.765\pmsd{.010} & 0.582\pmsd{.034} & 0.730\pmsd{.011} & 0.713\pmsd{.059} & 4.7\pmsd{0.8} & 0.495\pmsd{.014} & 0.673\pmsd{.013} \\
& Shushant nepaliBERT (fixed tok.) & 0.721\pmsd{.012} & 0.752\pmsd{.013} & 0.581\pmsd{.007} & 0.715\pmsd{.011} & 0.680\pmsd{.020} & 6.1\pmsd{0.6} & 0.496\pmsd{.014} & 0.664\pmsd{.011} \\
\midrule
\multirow{3}{*}{\shortstack[l]{Commercial\\LLMs}}
& Claude Opus 5.5 & \textbf{0.896} & \textbf{0.891} & \textbf{0.838} & 0.862 & \textbf{0.903} & \textbf{0.4} & \textbf{0.750} & 0.824 \\
& Gemini 3.8 Flash & 0.891 & 0.887 & 0.786 & 0.795 & 0.895 & 1.1 & 0.658 & 0.744 \\
& Gemini 3.1 Pro & 0.882 & 0.887 & 0.705 & 0.770 & 0.878 & 0.6 & 0.563 & 0.707 \\
\midrule
\multirow{2}{*}{\shortstack[l]{Open-weight\\LLMs}}
& Qwen3.5-397B-A17B & 0.867 & 0.856 & 0.681 & 0.745 & 0.855 & 0.9 & 0.552 & 0.680 \\
& Qwen3.5-9B & 0.790 & 0.804 & 0.622 & 0.674 & 0.761 & 3.4 & 0.478 & 0.597 \\
\bottomrule
\end{tabular}
\end{table*}

\subsection{Three classes: encoders and LLMs are equal}
\label{sec:main}
Fine-tuned encoders get much higher scores than the TF-IDF baselines (Table~\ref{tab:main}). On three classes, NepBERTa with the fixed tokenizer gets a macro-F1 of 0.877 and IndicBERTv2 0.867, against 0.766 for the best TF-IDF model. IndicBERTv2 has the best development macro-F1 (0.896). The difference between the two encoders is not significant (+0.010, 95\% CI [$-0.012$, 0.032]).

The three commercial LLMs get 0.882 to 0.896 with no training data. Claude Opus 5.5 has the highest score, but its gain over NepBERTa is not significant (+0.019, CI [$-0.015$, 0.051]). The open-weight Qwen3.5-397B-A17B (0.867) is equal to IndicBERTv2, and the small Qwen3.5-9B (0.790) is near mBERT. The LLMs answer ``don't know'' or give no label on 0.6\% (Claude) to 3.3\% (Qwen3.5-397B-A17B) of test sentences. On all 3,189 sentences, the three-class LLM scores change by at most 0.026, so the test set is representative.

\subsection{Five points: LLMs are better at intensity}
On the five-point scale, the order changes. Claude Opus 5.5 gets a lenient macro-F1 of 0.838, against 0.721 for NepBERTa and 0.734 for IndicBERTv2. Its gain over NepBERTa is significant (+0.116, CI [0.036, 0.187]), and so is its gain over IndicBERTv2 (+0.104, CI [0.002, 0.186]). Claude also makes the fewest severe errors (0.4\%). Gemini 3.8 Flash (0.786) is second, but its gain over NepBERTa is not significant (CI [$-0.011$, 0.131]). The strict scores give the same order: 0.750 for Claude and at most 0.640 for an encoder.

The extreme classes limit the encoders. The training set has 14 sentences with only SNeg and 43 with only SPos. Over all encoders, the mean F1 on SNeg is 0 to 0.22, and MuRIL gets an F1 of 0 on SPos. Development macro-F1 is not a reliable guide on five points, because the development set has one sentence with only SNeg. It selects Rajan's NepaliBERT (0.795 on development), which is ninth on test. Thus we give all encoders and do not select one.

\subsection{Tokenizer settings must match pretraining}
The default tokenizers of the three Nepali BERTs remove 44.5\% to 44.6\% of all Devanagari combining marks. Only 35.1\% to 36.8\% of words get to the model intact, and fertility increases from 1.20--1.27 to 1.95--2.04 subwords per word.

\begin{table}[t]
\centering
\caption{Default and fixed tokenizer for the three Nepali BERTs. Fert.: subwords per word. Lost: combining marks deleted. MLM: top-1 probe accuracy. F1: test macro-F1 (mean of 3 seeds) on our three classes, NepCOV19Tweets and news topics. Bold: the better setting where the 95\% CI of the difference excludes zero.}
\label{tab:tok}
\setlength{\tabcolsep}{3.6pt}
\begin{tabular}{llcccccc}
\toprule
Model & Tok. & Fert. & Lost & MLM & Ours & COV19 & News \\
\midrule
\multirow{2}{*}{NepBERTa} & default & 2.04 & 44.5\% & 0.205 & 0.793 & 0.702 & 0.957 \\
 & fixed & 1.27 & 0.0\% & \textbf{0.518} & \textbf{0.877} & \textbf{0.730} & \textbf{0.981} \\
\midrule
\multirow{2}{*}{Rajan} & default & 1.95 & 44.6\% & 0.051$^\ast$ & 0.739 & 0.666 & 0.945 \\
 & fixed & 1.20 & 0.05\% & \textbf{0.176}$^\ast$ & \textbf{0.788} & 0.672 & \textbf{0.974} \\
\midrule
\multirow{2}{*}{Shushant} & default & 2.02 & 44.5\% & \textbf{0.650} & \textbf{0.817} & \textbf{0.700} & \textbf{0.975} \\
 & fixed & 1.26 & 0.0\% & 0.071 & 0.721 & 0.668 & 0.918 \\
\bottomrule
\multicolumn{8}{l}{\footnotesize $^\ast$The checkpoint has no MLM output bias; the values are not reliable.}
\end{tabular}
\end{table}

\begin{figure}[t]
\centering
\includegraphics[width=0.85\columnwidth]{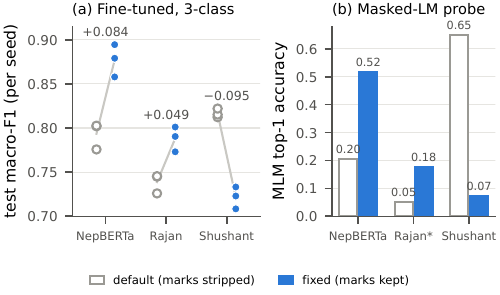}
\caption{(a) Three-class test macro-F1 of each seed with the default (hollow) and fixed (filled) tokenizer. The numbers give the mean change. (b) Accuracy of the MLM probe for each checkpoint. It predicts the direction of each effect.}
\label{fig:tok}
\end{figure}

The fix has a different effect on each checkpoint (Fig.~\ref{fig:tok}a). On three classes, it raises the macro-F1 of NepBERTa by 0.084 (95\% CI [0.057, 0.115]) and of Rajan's NepaliBERT by 0.049 (CI [0.019, 0.079]). It lowers the macro-F1 of Shushant's nepaliBERT by 0.095 (CI [$-0.131$, $-0.063$]). All three effects have the same direction on all seeds. On five points, the changes are $+0.050$, $+0.036$ and $-0.089$, and all CIs exclude zero.

The MLM probe predicts each direction (Fig.~\ref{fig:tok}b). NepBERTa predicts masked tokens much better with the marks (top-1 accuracy 0.518 against 0.205), so its pretraining text had the marks. The Shushant checkpoint shows the opposite (0.650 stripped, 0.071 kept), so its pretraining text was stripped. Thus the fixed tokenizer gives it input that it did not see in pretraining. The probe needs no labels, so it is a quick check before fine-tuning.

On two public datasets, NepCOV19Tweets~\cite{sitaula2021deep} and the Nepali news topics of MTEB, the fix has the same effect (Table~\ref{tab:tok}). It helps NepBERTa and hurts Shushant on all seeds, and the four CIs exclude zero. Rajan shows a gain on news but not on NepCOV19Tweets.

\subsection{Loss, training size and errors}
An ordinal loss gives no consistent gain on five points. Cross-entropy (CE) plus the expected distance $\sum_k p_k\,|k-y|$ raises the lenient macro-F1 of NepBERTa from 0.721 to 0.746, which is less than one seed standard deviation. It lowers that of IndicBERTv2 from 0.734 to 0.691. With 10\% of the training data (224 sentences), NepBERTa gets 0.833\pmsd{.014} on three classes, which is more than TF-IDF on all data (0.766). IndicBERTv2 is unstable with 10\% of the data (0.481\pmsd{.058}), but it gets 0.838 with 25\%.

Severe errors are rare and often come from sarcasm. NepBERTa makes the same severe error on all three seeds for only two test sentences. One is \dv{16} \dv{17} \dv{18} \dv{19} \dv{20} \dv{21} \dv{22} \dv{23} \dv{24} \dv{25} \dv{26} \dv{27} \dv{23} \dv{28} \dv{29} \dv{30} \dv{31} \dv{32} \dv{33} \dv{30} \dv{34} \dv{18} \dv{35} \dv{36} \dv{37} \dv{38} \dv{39} \dv{40} \dv{6} (``The victim girls died without justice because of PM Khadga Prasad Oli, so I could not give Children's Day wishes. I want to give these wishes to the PM himself, please accept.''). Both annotators say Neg, but the words of greeting make NepBERTa predict Pos.

\section{Discussion and Limitations}

\textbf{Agreed labels make a better reference.} Two blind annotators rarely disagree on the direction of sentiment, but often disagree on Neu against a polar label and on intensity. Thus we keep only the agreed polarity, and we accept two labels where the annotators differ only in intensity. We recommend three reports for each new model: three-class, lenient five-point and strict five-point scores.

\textbf{LLMs are now competitive.} On three classes, the best commercial LLMs equal the best fine-tuned encoders with no training data. On five points, Claude Opus 5.5 is better than every encoder, because the encoders see few extreme examples. The open-weight Qwen3.5-397B-A17B equals a mid-sized encoder on three classes.

\textbf{Check the tokenizer before fine-tuning.} Make sure that the tokenizer keeps the script, and that it matches the normalization of the checkpoint. A blind fix would cost the Shushant model 0.095 macro-F1, and the default would cost NepBERTa 0.084.

\textbf{Limitations.} The agreed subset is easier than the full dataset, because we removed the 1,704 sentences where the annotators disagree on polarity. Thus our scores are higher than scores on all sentences would be, and the benchmark does not test the Neu--Pos boundary well. Only two annotators labelled the benchmark. The SNeg and SPos classes have few sentences with one label. We used three seeds, one set of hyperparameters and one zero-shot prompt for each LLM, and Claude saw its sentences in batches.

\section{Conclusion}

We built a Nepali sentiment benchmark from the agreed labels of two blind annotators. On three classes, NepBERTa with a fixed tokenizer, IndicBERTv2 and three commercial LLMs score about 0.87 to 0.90 macro-F1. On five points, Claude Opus 5.5 is best by a significant margin. Next, we will add annotators to the sentences with disagreement and release them as a second test set.

\bibliographystyle{IEEEtran}
\bibliography{refs}

\end{document}